\documentclass[11pt,lettersize]{article}
\usepackage[utf8]{inputenc}
\usepackage{amsmath,amssymb,amsthm,bm}
\usepackage{fullpage}
\usepackage{graphicx}
\usepackage{subcaption}

\begin{document}
\title{Interpreting Deep Learning: The Machine Learning Rorschach Test?}
\date{}
\author{Adam S. Charles\footnote{A.S.C. is with the Princeton Neuroscience Department at Princeton University, Princeton, NJ. 08544 (contact: adamsc@princeton.edu)}}
\maketitle

% \section{Introduction}

Theoretical understanding of deep learning is one of the most important tasks facing the statistics and machine learning communities. 
While multilayer, or deep, neural networks (DNNs) originated as models of biological networks in neuroscience~\cite{mcculloch1943logical,hopfield1982neural,hopfield1985neural,grossberg1988nonlinear} and psychology~\cite{levine1983neural,rumelhart1987parallel}, and as engineering methods~\cite{pham1970neural,ersu1984application}, they have become a centerpiece of the machine learning (ML) toolbox.
% Neural networks have moved on from their roots as models of biological networks in neuroscience~\cite{mcculloch1943logical,hopfield1982neural,hopfield1985neural,grossberg1988nonlinear} and in psychology~\cite{levine1983neural,rumelhart1987parallel}, and as engineering methods~\cite{pham1970neural,ersu1984application} to become a centerpiece of the machine learning toolbox. 
%Their success has spurred other fields with complex data analysis needs to adopt these methods. The ease of collecting large datasets and the open sharing of reliable, stable code for training neural networks has resulted in widespread adoption of these methods faster than the statistics and machine learning communities could fully analyze how and when these networks work or fail. For example, a full understanding of when and why so-called adversarial examples~\cite{szegedy2013intriguing,nguyen2015deep,moosavi2016deepfool,brown2017adversarial} exist is still missing, potentially leaving many neural network-based systems open to malicious activity. This gap in understanding has recently drawn attention of a number of researchers from a diverse set of backgrounds, who have begun to take a formal, theoretical approach to understanding artificial neural networks, with an emphasis on the most prevalent of such methods: deep neural networks (DNNs). 
In ML, DNNs are simultaneously one of the simplest and most complex methods. They consist of many interconnected nodes that are grouped into layers (see Figure~1a), whose operations are stunningly simple; the $n^{th}$ node of the network at a given layer $i$, $x_i(n)$ is simply a nonlinear function $f(\cdot)$ (e.g. saturating nonlinearity) applied to an affine function of the previous layer 
\begin{gather}
    x_i(n) = f\left( \bm{w}_i(n)\bm{x}_{i-1} +  b_i(n) \right), \nonumber
\end{gather}
where $\bm{x}_{i-1}\in\mathbb{R}^{N_i}$ is the network node values at the previous layer, $\bm{w}_i(n)\in\mathbb{R}^{N_i}$ is the linear weight matrix that projects the previous layer to the $n^{th}$ node of the current matrix and $b_i(n)$ is the offset for node $n$. Even with such simple functions connecting the nodes between layers, the sheer number of nodes creates an explosion in the number of parameters ($\bm{w}_i(n)$ and $b_i(n)$ for all $i$ and $n$) and amplifies the effects of the nonlinearities. To add to the complexity, the parameters of the network (i.e. the weights and offsets across layers) are learned with respect to a cost function relating the inputs and outputs by gradient descent methods, i.e. various flavors of back-propagation~\cite{rumelhart1986learning}. Despite the resulting complexity, researchers have utilized DNNs to great effect in many important applications. 

The relatively recent success of DNNs in ML, despite their long history, can be attributed to a ``perfect storm'' of large labeled datasets~\cite{deng2012mnist,deng2009imagenet}; improved hardware~\cite{jouppi2017datacenter}; clever parameter constraints~\cite{krizhevsky2012imagenet}; advancements in optimization algorithms~\cite{kingma2014adam,sutskever2013importance,johnson2013accelerating}; and more open sharing of stable, reliable code~\cite{abadi2016tensorflow} leveraging the latest in methods such as automatic differentiation~\cite{rall1981automatic}. Original tasks in which DNNs first provided state-of-the-art results centered around image classification~\cite{krizhevsky2012imagenet,lecun1998gradient}, which powers devices such as ATMs. While DNNs have spread well beyond to many other applications (e.g. audio classification~\cite{hinton2012deep}, probability distribution approximation~\cite{kingma2013auto,makhzani2015adversarial} etc.),
the well publicized success in image classification has encouraged continued work that has provided other amazing technologies such as real-time text translation~\cite{good2015blog}. 

Unfortunately, DNN adoption powered by these successes combined with the open-source nature of the machine learning community, has outpaced our theoretical understanding. We cannot reliably identify when and why DNNs will make mistakes. In some applications like text translation these mistakes may be comical and provide for fun fodder in research talks, a single error can be very costly in tasks like medical imaging~\cite{finlayson2018adversarial}. 
Additionally, DNNs shown susceptibility to so-called adversarial examples, or data specifically designed to fool a DNN~\cite{szegedy2013intriguing,nguyen2015deep,moosavi2016deepfool,brown2017adversarial}. One can generate such examples with imperceptible deviations from an image, causing the system to mis-classify an image that is nearly identical to a correctly classified one. Audio adversarial examples can also exert control over popular systems such as Amazon Alexa or Siri, allowing malicious access to devices containing personal information~\cite{carlini2018audio,zhang2017dolphinattack}. As we utilize DNNs in increasingly sensitive applications, a better understanding of their properties is thus imperative. 

Early theory of DNNs or multi-layered networks (the smooth-nonlinearity versions of the non-smooth multi-layered perceptrons~\cite{minsky1990perceptrons}) were thought of more generally as learning machines and early theory sought to use statistical learning theory~\cite{valiant1984theory,vapnik1998statistical,cucker2002mathematical} or function approximation theory~\cite{hornik1991approximation} to analyze quantities such as the Vapnik-Chervonenkis (VC) dimension of DNNs~\cite{vapnik1994measuring}. While these theories address generally the complexity of neural networks with respect to training data, many important questions pertaining to the expressibility, learning rule efficiency, intuition, susceptibility to adversarial examples etc.\ remain. 

Recently, a number of theories spanning subsets of these questions have been proposed. These analyses mostly fall into three main styles of analysis. First are the methods that aim to show how DNNs perform explicit mathematical tasks by demonstrating how specific combinations of nonlinearities and weights recover exactly a known (and typically general) function on the data. The second method tries instead to describe the theoretical limitations and capabilities of the sequence of functions present in any DNN, again typically with constraints or assumptions about the nonlinearities and weights. These analyses can also be functions of the data, for example analyses that try to quantify and understand the cost-function landscape, which depends intimately on the data used for training (e.g.~\cite{ballard2017energy,mhaskar2018analysis}). The third area of DNN theory that is worthy of note is the literature analyzing the abilities of specific algorithms to efficiently solve the high-dimensional, nonlinear optimization programs required to train DNNs (e.g.~\cite{wilson2017marginal,zhang2018theory}). These analyses focus on the interplay between the training algorithm and the properties of DNNs as a mathematical structure. 

Advances in analyzing DNNs have included many different sources of intuition, drawn on both observations about the connections between the local and global computations DNNs perform to operations from other fields as well as applications of various analysis methods to understand how these operations interact. For example, the iterative linear-then-threshold structure has been related to the steps needed to find sparse representations of images~\cite{borgerding2017amp,xin2016maximal,papyan2016convolutional}. This result draws connections to temporally un-rolled iterative algorithms that explicitly solve penalized optimization programs such as basis pursuit de-noising (BPDN, aka LASSO). Specifically, solving the regularized least-squares optimization
\begin{gather}
\arg\min_{\bm{\beta}} \left[ \left\|\bm{y} - \bm{A}\bm{\beta} \right\| + \lambda R(\bm{\beta})\right],
\end{gather}
via a proximal projection method amounts to iteratively calculating
\begin{gather}
\widehat{\bm{\beta}}_{t+1} = P_{\lambda}\left(\widehat{\bm{\beta}}_{t} + \bm{A}^T\left(\bm{y} - \bm{A}\widehat{\bm{\beta}}_{t}) \right)\right),
\end{gather}
where $P_{\lambda}(\bm{z})$ is the nonlinearity that calculates the proximal projection $\min_{\beta} \|\bm{z} -\bm{\beta}\|_2^2 + \lambda R(\bm{\beta})$. In the case where the regularization function $R(\cdot)$ is separable, the proximal projection is a point-wise nonlinearity, mimicking the form of DNNs. Treating $\widehat{\bm{\beta}}_{t}$ at each algorithmic iteration as a different set of variables, these variables can be considered the node values at different layers of a deep neural network with weights $\bm{A}^T\bm{A} + \bm{I}$ between layers, a bias $\bm{A}^T\bm{y}$ at each layer the nonlinearity defined by the proximal projection. This example gives a flavor of how the network weights and nonlinearity can be mapped to a specific operation to understand the functionality of DNNs. A more global computational interpretation looks at the computation of all layers as a whole, drawing a connection to tensor decompositions and classification based on tensor inner products~\cite{cohen2016convolutional,cohen2017analysis}. A non-exhaustive list of additional interpretations and analysis thus far includes:
\begin{itemize}
    \item Complexity analysis such as Chaitin-Kolmogorov complexity~\cite{pearlmutter1991chaitin}, Vapnik-Chervonenkis (VC) dimension calculations~\cite{bartlett1993lower,bartlett1993vapnik,vapnik1994measuring,bartlett1996vc,bartlett1999almost,bartlett2003vapnik}, sample-complexity analysis~\cite{bartlett1998sample}, Lipshitz-based generalization bounds~\cite{bartlett2017spectrally}
    \item analysis of the ability of deep networks as function approximators~\cite{hornik1991approximation,shaham2016provable,mhaskar2016deep,eldan2016power,long2018representing}
    \item inspirations from physics (including low-dimensional structure~\cite{lin2017does}, renormalization~\cite{mehta2014exact}, tools from high-dimensional statistical mechanics~\cite{poole2016exponential}, quantum entanglement~\cite{levine2017deep,levine2018bridging} and connections to the information bottleneck~\cite{tishby2015deep})
    \item chemistry (in interpreting the energy landscape~\cite{ballard2017energy})
    \item connections to wavelet transforms and invariance~\cite{mallat2016understanding,bruna2013invariant,wiatowski2016discrete}
    \item connections to message passing algorithms that marginalize nuisance variables~\cite{patel2016probabilistic}
    \item generalization analysis via cost-function maxima~\cite{mhaskar2018analysis}
    \item high-dimensional probability theory for analyzing the network Jacobaian~\cite{pennington2017resurrecting,pennington2018emergence} or layer-wise Gramm matrix~\cite{pennington2017nonlinear}
    \item equivalence to Gaussian processes in certain limiting cases~\cite{lee2017deep}
    \item equivalence of DNNs to hierarchical tensor decompositions~\cite{cohen2016convolutional,cohen2017analysis,stock2018learning}
    \item relations to better understood single-layer networks~\cite{veit2016residual,philipp2017gradients} 
    \item analysis of the invertability and information retention through CNN layers~\cite{arora2015deep,bahmani2017anchored}
    \item complexity analysis on the learnability of neural networks~\cite{song2017complexity} 
    \item algebraic topology approaches to understanding the complexity of data and choosing DNN architectures~\cite{guss2018characterizing}
    \item probing DNN functionality using tools from psychology~\cite{ritter2017cognitive}
    \item empirical analysis via influence functions~\cite{koh2017understanding}
    \item analysis of specific counter-examples~\cite{safran2017depth,safran2017spurious}
    \item DNN compression-based generalization analysis~\cite{arora2018stronger}
    \item interpretations as hierarchical kernel machines~\cite{anselmi2015deep}
    \item information theory~\cite{yu2018understanding}
    \item analysis of the role of layered representations on DNN properties via the learning dynamics~\cite{haeffele2015global,sagun2017empirical,shalev2017failures,safran2016quality,saxe2013exact}
    \item speed and accuracy guarantees (or lack thereof) of learning methods for DNNs~\cite{arora2014provable,pascanu2014saddle,dauphin2014identifying,arora2016provable,wilson2017marginal,arora2018optimization,kidambi2018insufficiency}  
\end{itemize}

This list, which still omits specialized analyses of specific optimization tricks such as dropout~\cite{gal2015dropout} and newer architectures such as generative adversarial networks (GANS)~\cite{arora2017generalization,arora2017gans} or deep recurrent networks~\cite{levine2017benefits}, demonstrates just how relentless the search for meaning in DNNs has become. In the breadth of possible interpretations, some interesting points begin to emerge. For one, there seems to be a limitless number of interpretations for DNNs, apparently constrained only by the lens by which the mathematical operations are viewed. Physics interpretations stem from researchers with a physics background. Connections to sparsity and wavelets come from researchers well known for important contributions to those fields. 
Ultimately, the interpretation of DNNs appears to mimic a type of Rorschach test --- a psychological test wherein subjects interpret a series of seemingly ambiguous ink-blots~\cite{rorschach1922psychodiagnostik} (see Figure~1). Rorschach tests depend not only on \emph{what} (the result) a subject sees in the ink-blots but also on the \emph{reasoning} (methods used) behind the subject's perception, thus making the analogy particularly apropos. 

\begin{figure*}[h]
\centering
    \begin{subfigure}[t]{0.5\textwidth}
        \centering
        \includegraphics[width=0.9\textwidth]{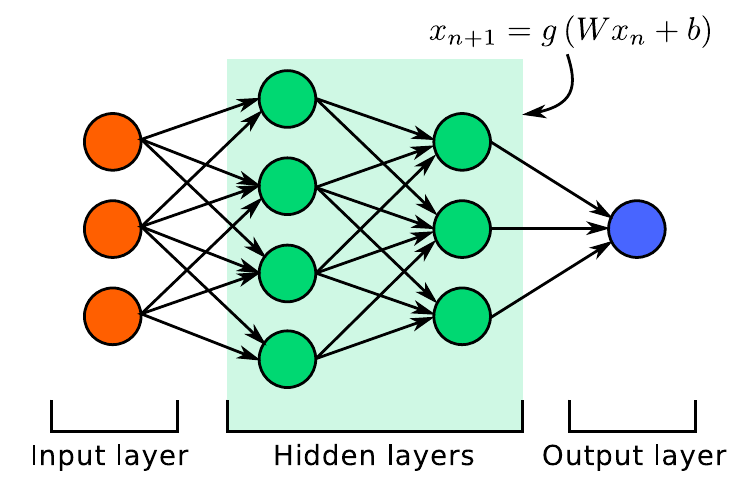}
    \end{subfigure}%
    ~ 
    \begin{subfigure}[t]{0.5\textwidth}
        \centering
        \includegraphics[width=0.9\textwidth]{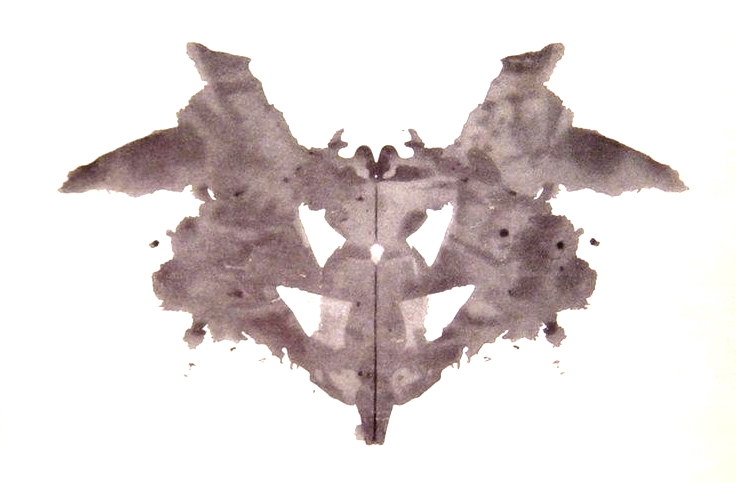}
    \end{subfigure}
        \caption{What do you see? DNNs can be viewed in many ways. 1a. Stylistic example of a DNN with an input layer (red), output layer (blue) and two hidden layers (green); example ``ink blot'' for DNN theory. 1b. Example (normalized) ink blot from the Rorschach test.}
\end{figure*}

On the one hand, is unsurprising given DNNs status as arbitrary function approximators. Specific network weights and nonlinearities allow DNNs to easily adapt to various narratives. On the other hand, they are not unique in their permitting multiple interpretations. One can likewise view standard, simpler, algorithms through various lenses. For example one can derive the Kalman filter --- a time-tested algorithm for tracking a vector over time --- from at least three interpretations: : the orthogonality principle~\cite{haykin2008adaptive}, Bayesian maximum \emph{a-priori} estimation~\cite{barker1995bayesian,charles2015dynamics}, and low-rank updates for least-squares optimization~\cite{moon2000mathematical}. These three derivations allow people with different mathematical mindsets (i.e., linear algebra versus probability theory) to understand the algorithm.

Yet compared to DNNs, the Kalman filter is simple, consisting of only a handful of linear-algebraic operations. It's function is completely understood, allowing each viewpoint to be validated despite the different underlying philosophies.  
Similar validation for DNN theory requires a convergence of the literature. We must distinguish between universal results that are invariant to the analysis perspective and those that are specific to a particular network configuration. A healthy debate is already underway, with respect to the information bottleneck interpretation of DNNs~\cite{tishby2015deep,saxe2018iclr}. We must also better understand how the functions DNNs perform, their mathematical properties, and the impact of the optimization methods interact. The complexity of DNNs, however, introduces many challenges. For one, many standard tools (e.g.\ for understanding how models generalize from training data~\cite{zhang2016understanding} or empirically assessing important network features~\cite{ghorbani2017interpretation}) are difficult to apply to DNNs.
Luckily, there is no shortage of excitement, and we continue to enhance our understanding of DNNs with time.  
The community is also beginning to coalesce, and dedicated meetings like recent workshops at the Conference on Neural Information Processing Systems (NIPS) and the recent Mathematical Theory of Deep Neural Network symposium at Princeton University, will further accelerate our pace. 

Additionally, it is worth mentioning that a similar branching of neural network analysis --- separating out what functions they can solve and the generic properties they have --- has occurred for recurrent neural networks (RNNs)~\cite{hopfield1982neural} as well. Separate literatures have evolved to analyze the functions (typically as an optimization functions) RNNs can solve~\cite{rozell2008sparse,hu2012network,charles2012common} and what generic properties such networks can have (e.g.\ the low-dimensional dynamics~\cite{sussillo2013opening,rivkind2017local} eigenvalue structure~\cite{rajan2006eigenvalue}, expressiveness~\cite{khrulkov2017expressive} or short-term memory of RNNs~\cite{jaeger2001short,buonomano2009state,charles2014short,maass2002real,charles2017distributed}). These questions are especially pertinent given the current rise in the use of RNNs in fields such as neuroscience (e.g.~\cite{zhu2013visual,sussillo2015neural,depasquale2018full,rajan2016recurrent}). Like the DNN literature, a bridging between these two distinct ways of analyzing the same mathematical object has not yet taken place and will similarly be key to substantial progress in our understanding of these systems. 

%Bridging the gap between the two main approaches to analyzing DNNs (and likewise for RNNs) will possibly be key to overcoming some of the big challenges in finding commonality between all the varying viewpoints, and the neural network literature will have much to gain in making this connection.

\vspace{10pt}

{\bf Acknowledgements:} I would like to thank Justin Romberg, Jonathan Pillow, Michael Shvartsman, Mikio Aoi, Ahmed El Hady Camille Rullan, David Zoltowski, and the editorial staff of \emph{SIAM News} (Lina Sorg and Karthika Swamy Cohen), all of who have kindly provided feedback on this manuscript and helped me refine its ideas and message.

\bibliographystyle{unsrt}
\bibliography{rt.bib}

\begin{thebibliography}{100}

\bibitem{mcculloch1943logical}
Warren~S McCulloch and Walter Pitts.
\newblock A logical calculus of the ideas immanent in nervous activity.
\newblock {\em The bulletin of mathematical biophysics}, 5(4):115--133, 1943.

\bibitem{hopfield1982neural}
John~J Hopfield.
\newblock Neural networks and physical systems with emergent collective
  computational abilities.
\newblock {\em Proceedings of the national academy of sciences},
  79(8):2554--2558, 1982.

\bibitem{hopfield1985neural}
John~J Hopfield and David~W Tank.
\newblock “neural” computation of decisions in optimization problems.
\newblock {\em Biological cybernetics}, 52(3):141--152, 1985.

\bibitem{grossberg1988nonlinear}
Stephen Grossberg.
\newblock Nonlinear neural networks: Principles, mechanisms, and architectures.
\newblock {\em Neural networks}, 1(1):17--61, 1988.

\bibitem{levine1983neural}
Daniel~S Levine.
\newblock Neural population modeling and psychology: A review.
\newblock {\em Mathematical Biosciences}, 66(1):1--86, 1983.

\bibitem{rumelhart1987parallel}
David~E Rumelhart, James~L McClelland, PDP~Research Group, et~al.
\newblock {\em Parallel distributed processing}, volume~1.
\newblock MIT press Cambridge, MA, 1987.

\bibitem{pham1970neural}
DT~Pham.
\newblock Neural networks in engineering.
\newblock {\em WIT Transactions on Information and Communication Technologies},
  6, 1970.

\bibitem{ersu1984application}
E~Ers{\"u}.
\newblock On the application of associative neural network models to technical
  control problems.
\newblock In {\em Localization and Orientation in Biology and Engineering},
  pages 90--93. Springer, 1984.

\bibitem{rumelhart1986learning}
David~E Rumelhart, Geoffrey~E Hinton, and Ronald~J Williams.
\newblock Learning representations by back-propagating errors.
\newblock {\em nature}, 323(6088):533, 1986.

\bibitem{deng2012mnist}
Li~Deng.
\newblock The mnist database of handwritten digit images for machine learning
  research [best of the web].
\newblock {\em IEEE Signal Processing Magazine}, 29(6):141--142, 2012.

\bibitem{deng2009imagenet}
Jia Deng, Wei Dong, Richard Socher, Li-Jia Li, Kai Li, and Li~Fei-Fei.
\newblock Imagenet: A large-scale hierarchical image database.
\newblock In {\em Computer Vision and Pattern Recognition, 2009. CVPR 2009.
  IEEE Conference on}, pages 248--255, 2009.

\bibitem{jouppi2017datacenter}
Norman~P Jouppi, Cliff Young, Nishant Patil, David Patterson, Gaurav Agrawal,
  Raminder Bajwa, Sarah Bates, Suresh Bhatia, Nan Boden, Al~Borchers, et~al.
\newblock In-datacenter performance analysis of a tensor processing unit.
\newblock In {\em Proceedings of the 44th Annual International Symposium on
  Computer Architecture}, pages 1--12, 2017.

\bibitem{krizhevsky2012imagenet}
Alex Krizhevsky, Ilya Sutskever, and Geoffrey~E Hinton.
\newblock Imagenet classification with deep convolutional neural networks.
\newblock In {\em Advances in neural information processing systems}, pages
  1097--1105, 2012.

\bibitem{kingma2014adam}
Diederik~P Kingma and Jimmy Ba.
\newblock Adam: A method for stochastic optimization.
\newblock {\em arXiv preprint arXiv:1412.6980}, 2014.

\bibitem{sutskever2013importance}
Ilya Sutskever, James Martens, George Dahl, and Geoffrey Hinton.
\newblock On the importance of initialization and momentum in deep learning.
\newblock In {\em International conference on machine learning}, pages
  1139--1147, 2013.

\bibitem{johnson2013accelerating}
Rie Johnson and Tong Zhang.
\newblock Accelerating stochastic gradient descent using predictive variance
  reduction.
\newblock In {\em Advances in neural information processing systems}, pages
  315--323, 2013.

\bibitem{abadi2016tensorflow}
Mart{\'\i}n Abadi, Paul Barham, Jianmin Chen, Zhifeng Chen, Andy Davis, Jeffrey
  Dean, Matthieu Devin, Sanjay Ghemawat, Geoffrey Irving, Michael Isard, et~al.
\newblock Tensorflow: A system for large-scale machine learning.
\newblock In {\em OSDI}, volume~16, pages 265--283, 2016.

\bibitem{rall1981automatic}
Louis~B Rall.
\newblock Automatic differentiation: Techniques and applications.
\newblock 1981.

\bibitem{lecun1998gradient}
Yann LeCun, L{\'e}on Bottou, Yoshua Bengio, and Patrick Haffner.
\newblock Gradient-based learning applied to document recognition.
\newblock {\em Proceedings of the IEEE}, 86(11):2278--2324, 1998.

\bibitem{hinton2012deep}
Geoffrey Hinton, Li~Deng, Dong Yu, George~E Dahl, Abdel-rahman Mohamed, Navdeep
  Jaitly, Andrew Senior, Vincent Vanhoucke, Patrick Nguyen, Tara~N Sainath,
  et~al.
\newblock Deep neural networks for acoustic modeling in speech recognition: The
  shared views of four research groups.
\newblock {\em IEEE Signal Processing Magazine}, 29(6):82--97, 2012.

\bibitem{kingma2013auto}
Diederik~P Kingma and Max Welling.
\newblock Auto-encoding variational bayes.
\newblock {\em arXiv preprint arXiv:1312.6114}, 2013.

\bibitem{makhzani2015adversarial}
Alireza Makhzani, Jonathon Shlens, Navdeep Jaitly, Ian Goodfellow, and Brendan
  Frey.
\newblock Adversarial autoencoders.
\newblock {\em arXiv preprint arXiv:1511.05644}, 2015.

\bibitem{good2015blog}
Otavio Good.
\newblock How google translate squeezes deep learning onto a phone.
\newblock
  https://ai.googleblog.com/2015/07/how-google-translate-squeezes-deep.html,
  2015.

\bibitem{finlayson2018adversarial}
Samuel~G Finlayson, Isaac~S Kohane, and Andrew~L Beam.
\newblock Adversarial attacks against medical deep learning systems.
\newblock {\em arXiv preprint arXiv:1804.05296}, 2018.

\bibitem{szegedy2013intriguing}
Christian Szegedy, Wojciech Zaremba, Ilya Sutskever, Joan Bruna, Dumitru Erhan,
  Ian Goodfellow, and Rob Fergus.
\newblock Intriguing properties of neural networks.
\newblock {\em arXiv preprint arXiv:1312.6199}, 2013.

\bibitem{nguyen2015deep}
Anh Nguyen, Jason Yosinski, and Jeff Clune.
\newblock Deep neural networks are easily fooled: High confidence predictions
  for unrecognizable images.
\newblock In {\em Proceedings of the IEEE Conference on Computer Vision and
  Pattern Recognition}, pages 427--436, 2015.

\bibitem{moosavi2016deepfool}
Seyed~Mohsen Moosavi~Dezfooli, Alhussein Fawzi, and Pascal Frossard.
\newblock Deepfool: a simple and accurate method to fool deep neural networks.
\newblock In {\em Proceedings of 2016 IEEE Conference on Computer Vision and
  Pattern Recognition (CVPR)}, number EPFL-CONF-218057, 2016.

\bibitem{brown2017adversarial}
Tom~B Brown, Dandelion Man{\'e}, Aurko Roy, Mart{\'\i}n Abadi, and Justin
  Gilmer.
\newblock Adversarial patch.
\newblock {\em arXiv preprint arXiv:1712.09665}, 2017.

\bibitem{carlini2018audio}
Nicholas Carlini and David Wagner.
\newblock Audio adversarial examples: Targeted attacks on speech-to-text.
\newblock {\em arXiv preprint arXiv:1801.01944}, 2018.

\bibitem{zhang2017dolphinattack}
Guoming Zhang, Chen Yan, Xiaoyu Ji, Tianchen Zhang, Taimin Zhang, and Wenyuan
  Xu.
\newblock Dolphinattack: Inaudible voice commands.
\newblock In {\em Proceedings of the 2017 ACM SIGSAC Conference on Computer and
  Communications Security}, pages 103--117, 2017.

\bibitem{minsky1990perceptrons}
Marvin Minsky and Seymour Papert.
\newblock Perceptrons.
\newblock {\em MIT Press}, 1969.

\bibitem{valiant1984theory}
Leslie~G Valiant.
\newblock A theory of the learnable.
\newblock {\em Communications of the ACM}, 27(11):1134--1142, 1984.

\bibitem{vapnik1998statistical}
Vladimir Vapnik.
\newblock {\em Statistical learning theory. 1998}.
\newblock Wiley, New York, 1998.

\bibitem{cucker2002mathematical}
Felipe Cucker and Steve Smale.
\newblock On the mathematical foundations of learning.
\newblock {\em Bulletin of the American mathematical society}, 39(1):1--49,
  2002.

\bibitem{hornik1991approximation}
Kurt Hornik.
\newblock Approximation capabilities of multilayer feedforward networks.
\newblock {\em Neural networks}, 4(2):251--257, 1991.

\bibitem{vapnik1994measuring}
Vladimir Vapnik, Esther Levin, and Yann~Le Cun.
\newblock Measuring the vc-dimension of a learning machine.
\newblock {\em Neural computation}, 6(5):851--876, 1994.

\bibitem{ballard2017energy}
Andrew~J Ballard, Ritankar Das, Stefano Martiniani, Dhagash Mehta, Levent
  Sagun, Jacob~D Stevenson, and David~J Wales.
\newblock Energy landscapes for machine learning.
\newblock {\em Physical Chemistry Chemical Physics}, 19(20):12585--12603, 2017.

\bibitem{mhaskar2018analysis}
Hrushikesh Mhaskar and Tomaso Poggio.
\newblock An analysis of training and generalization errors in shallow and deep
  networks.
\newblock {\em arXiv preprint arXiv:1802.06266}, 2018.

\bibitem{wilson2017marginal}
Ashia~C Wilson, Rebecca Roelofs, Mitchell Stern, Nati Srebro, and Benjamin
  Recht.
\newblock The marginal value of adaptive gradient methods in machine learning.
\newblock In {\em Advances in Neural Information Processing Systems}, pages
  4151--4161, 2017.

\bibitem{zhang2018theory}
Chiyuan Zhang, Qianli Liao, Alexander Rakhlin, Brando Miranda, Noah Golowich,
  and Tomaso Poggio.
\newblock Theory of deep learning iib: Optimization properties of sgd.
\newblock {\em arXiv preprint arXiv:1801.02254}, 2018.

\bibitem{borgerding2017amp}
Mark Borgerding, Philip Schniter, and Sundeep Rangan.
\newblock Amp-inspired deep networks for sparse linear inverse problems.
\newblock {\em IEEE Transactions on Signal Processing}, 2017.

\bibitem{xin2016maximal}
Bo~Xin, Yizhou Wang, Wen Gao, David Wipf, and Baoyuan Wang.
\newblock Maximal sparsity with deep networks?
\newblock In {\em Advances in Neural Information Processing Systems}, pages
  4340--4348, 2016.

\bibitem{papyan2016convolutional}
Vardan Papyan, Yaniv Romano, and Michael Elad.
\newblock Convolutional neural networks analyzed via convolutional sparse
  coding.
\newblock {\em stat}, 1050:27, 2016.

\bibitem{cohen2016convolutional}
Nadav Cohen and Amnon Shashua.
\newblock Convolutional rectifier networks as generalized tensor
  decompositions.
\newblock In {\em International Conference on Machine Learning}, pages
  955--963, 2016.

\bibitem{cohen2017analysis}
Nadav Cohen, Or~Sharir, Yoav Levine, Ronen Tamari, David Yakira, and Amnon
  Shashua.
\newblock Analysis and design of convolutional networks via hierarchical tensor
  decompositions.
\newblock {\em arXiv preprint arXiv:1705.02302}, 2017.

\bibitem{pearlmutter1991chaitin}
Barak~A Pearlmutter and Ronald Rosenfeld.
\newblock Chaitin-kolmogorov complexity and generalization in neural networks.
\newblock In {\em Advances in neural information processing systems}, pages
  925--931, 1991.

\bibitem{bartlett1993lower}
Peter~L Bartlett.
\newblock Lower bounds on the vapnik-chervonenkis dimension of multi-layer
  threshold networks.
\newblock In {\em Proceedings of the sixth annual conference on Computational
  learning theory}, pages 144--150. ACM, 1993.

\bibitem{bartlett1993vapnik}
Peter~L Bartlett.
\newblock Vapnik-chervonenkis dimension bounds for two-and three-layer
  networks, 1993.

\bibitem{bartlett1996vc}
Peter~L Bartlett and Robert~C Williamson.
\newblock The vc dimension and pseudodimension of two-layer neural networks
  with discrete inputs.
\newblock {\em Neural computation}, 8(3):625--628, 1996.

\bibitem{bartlett1999almost}
Peter~L Bartlett, Vitaly Maiorov, and Ron Meir.
\newblock Almost linear vc dimension bounds for piecewise polynomial networks.
\newblock In {\em Advances in Neural Information Processing Systems}, pages
  190--196, 1999.

\bibitem{bartlett2003vapnik}
Peter~L Bartlett and Wolfgang Maass.
\newblock Vapnik-chervonenkis dimension of neural nets.
\newblock {\em The handbook of brain theory and neural networks}, pages
  1188--1192, 2003.

\bibitem{bartlett1998sample}
Peter~L Bartlett.
\newblock The sample complexity of pattern classification with neural networks:
  the size of the weights is more important than the size of the network.
\newblock {\em IEEE transactions on Information Theory}, 44(2):525--536, 1998.

\bibitem{bartlett2017spectrally}
Peter~L Bartlett, Dylan~J Foster, and Matus~J Telgarsky.
\newblock Spectrally-normalized margin bounds for neural networks.
\newblock In {\em Advances in Neural Information Processing Systems}, pages
  6241--6250, 2017.

\bibitem{shaham2016provable}
Uri Shaham, Alexander Cloninger, and Ronald~R Coifman.
\newblock Provable approximation properties for deep neural networks.
\newblock {\em Applied and Computational Harmonic Analysis}, 2016.

\bibitem{mhaskar2016deep}
Hrushikesh~N Mhaskar and Tomaso Poggio.
\newblock Deep vs. shallow networks: An approximation theory perspective.
\newblock {\em Analysis and Applications}, 14(06):829--848, 2016.

\bibitem{eldan2016power}
Ronen Eldan and Ohad Shamir.
\newblock The power of depth for feedforward neural networks.
\newblock In {\em Conference on Learning Theory}, pages 907--940, 2016.

\bibitem{long2018representing}
Peter~L. Bartlett, Steven~N. Evans, and Philip~M. Long.
\newblock Representing smooth functions as compositions of near-identity
  functions with implications for deep network optimization.
\newblock {\em arXiv preprint arXiv:1804.05012}, 2018.

\bibitem{lin2017does}
Henry~W Lin, Max Tegmark, and David Rolnick.
\newblock Why does deep and cheap learning work so well?
\newblock {\em Journal of Statistical Physics}, 168(6):1223--1247, 2017.

\bibitem{mehta2014exact}
Pankaj Mehta and David~J Schwab.
\newblock An exact mapping between the variational renormalization group and
  deep learning.
\newblock {\em arXiv preprint arXiv:1410.3831}, 2014.

\bibitem{poole2016exponential}
Ben Poole, Subhaneil Lahiri, Maithra Raghu, Jascha Sohl-Dickstein, and Surya
  Ganguli.
\newblock Exponential expressivity in deep neural networks through transient
  chaos.
\newblock In {\em Advances in neural information processing systems}, pages
  3360--3368, 2016.

\bibitem{levine2017deep}
Yoav Levine, David Yakira, Nadav Cohen, and Amnon Shashua.
\newblock Deep learning and quantum entanglement: Fundamental connections with
  implications to network design.
\newblock {\em CoRR, abs/1704.01552}, 2017.

\bibitem{levine2018bridging}
Yoav Levine, Or~Sharir, Nadav Cohen, and Amnon Shashua.
\newblock Bridging many-body quantum physics and deep learning via tensor
  networks.
\newblock {\em arXiv preprint arXiv:1803.09780}, 2018.

\bibitem{tishby2015deep}
Naftali Tishby and Noga Zaslavsky.
\newblock Deep learning and the information bottleneck principle.
\newblock In {\em IEEE Information Theory Workshop}, pages 1--5, 2015.

\bibitem{mallat2016understanding}
St{\'e}phane Mallat.
\newblock Understanding deep convolutional networks.
\newblock {\em Phil. Trans. R. Soc. A}, 374(2065):20150203, 2016.

\bibitem{bruna2013invariant}
Joan Bruna and St{\'e}phane Mallat.
\newblock Invariant scattering convolution networks.
\newblock {\em IEEE transactions on pattern analysis and machine intelligence},
  35(8):1872--1886, 2013.

\bibitem{wiatowski2016discrete}
Thomas Wiatowski, Michael Tschannen, Aleksandar Stanic, Philipp Grohs, and
  Helmut B{\"o}lcskei.
\newblock Discrete deep feature extraction: A theory and new architectures.
\newblock In {\em International Conference on Machine Learning}, pages
  2149--2158, 2016.

\bibitem{patel2016probabilistic}
Ankit~B Patel, Minh~Tan Nguyen, and Richard Baraniuk.
\newblock A probabilistic framework for deep learning.
\newblock In {\em Advances in Neural Information Processing Systems}, pages
  2558--2566, 2016.

\bibitem{pennington2017resurrecting}
Jeffrey Pennington, Samuel Schoenholz, and Surya Ganguli.
\newblock Resurrecting the sigmoid in deep learning through dynamical isometry:
  theory and practice.
\newblock In {\em Advances in neural information processing systems}, pages
  4788--4798, 2017.

\bibitem{pennington2018emergence}
Jeffrey Pennington, Samuel~S Schoenholz, and Surya Ganguli.
\newblock The emergence of spectral universality in deep networks.
\newblock {\em arXiv preprint arXiv:1802.09979}, 2018.

\bibitem{pennington2017nonlinear}
Jeffrey Pennington and Pratik Worah.
\newblock Nonlinear random matrix theory for deep learning.
\newblock In {\em Advances in Neural Information Processing Systems}, pages
  2634--2643, 2017.

\bibitem{lee2017deep}
Jaehoon Lee, Yasaman Bahri, Roman Novak, Samuel~S Schoenholz, Jeffrey
  Pennington, and Jascha Sohl-Dickstein.
\newblock Deep neural networks as gaussian processes.
\newblock {\em arXiv preprint arXiv:1711.00165}, 2017.

\bibitem{stock2018learning}
Christopher~H Stock, Alex~H Williams, Madhu~S Advani, Andrew~M Saxe, and Surya
  Ganguli.
\newblock Learning dynamics of deep networks admit low-rank tensor
  descriptions.
\newblock 2018.

\bibitem{veit2016residual}
Andreas Veit, Michael~J Wilber, and Serge Belongie.
\newblock Residual networks behave like ensembles of relatively shallow
  networks.
\newblock In {\em Advances in Neural Information Processing Systems}, pages
  550--558, 2016.

\bibitem{philipp2017gradients}
George Philipp, Dawn Song, and Jaime~G Carbonell.
\newblock Gradients explode-deep networks are shallow-resnet explained.
\newblock {\em arXiv preprint arXiv:1712.05577}, 2017.

\bibitem{arora2015deep}
Sanjeev Arora, Yingyu Liang, and Tengyu Ma.
\newblock Why are deep nets reversible: A simple theory, with implications for
  training.
\newblock {\em arXiv preprint arXiv:1511.05653}, 2015.

\bibitem{bahmani2017anchored}
Sohail Bahmani and Justin Romberg.
\newblock Anchored regression: Solving random convex equations via convex
  programming.
\newblock {\em arXiv preprint arXiv:1702.05327}, 2017.

\bibitem{song2017complexity}
Le~Song, Santosh Vempala, John Wilmes, and Bo~Xie.
\newblock On the complexity of learning neural networks.
\newblock In {\em Advances in Neural Information Processing Systems}, pages
  5520--5528, 2017.

\bibitem{guss2018characterizing}
William~H Guss and Ruslan Salakhutdinov.
\newblock On characterizing the capacity of neural networks using algebraic
  topology.
\newblock {\em arXiv preprint arXiv:1802.04443}, 2018.

\bibitem{ritter2017cognitive}
Samuel Ritter, David~GT Barrett, Adam Santoro, and Matt~M Botvinick.
\newblock Cognitive psychology for deep neural networks: A shape bias case
  study.
\newblock {\em arXiv preprint arXiv:1706.08606}, 2017.

\bibitem{koh2017understanding}
Pang~Wei Koh and Percy Liang.
\newblock Understanding black-box predictions via influence functions.
\newblock {\em arXiv preprint arXiv:1703.04730}, 2017.

\bibitem{safran2017depth}
Itay Safran and Ohad Shamir.
\newblock Depth-width tradeoffs in approximating natural functions with neural
  networks.
\newblock In {\em International Conference on Machine Learning}, pages
  2979--2987, 2017.

\bibitem{safran2017spurious}
Itay Safran and Ohad Shamir.
\newblock Spurious local minima are common in two-layer relu neural networks.
\newblock {\em arXiv preprint arXiv:1712.08968}, 2017.

\bibitem{arora2018stronger}
Sanjeev Arora, Rong Ge, Behnam Neyshabur, and Yi~Zhang.
\newblock Stronger generalization bounds for deep nets via a compression
  approach.
\newblock {\em arXiv preprint arXiv:1802.05296}, 2018.

\bibitem{anselmi2015deep}
Fabio Anselmi, Lorenzo Rosasco, Cheston Tan, and Tomaso Poggio.
\newblock Deep convolutional networks are hierarchical kernel machines.
\newblock {\em arXiv preprint arXiv:1508.01084}, 2015.

\bibitem{yu2018understanding}
Shujian Yu, Robert Jenssen, and Jose~C. Principe.
\newblock Understanding convolutional neural network training with information
  theory.
\newblock {\em arXiv preprint arXiv:1804.06537}, 2018.

\bibitem{haeffele2015global}
Benjamin~D Haeffele and Ren{\'e} Vidal.
\newblock Global optimality in tensor factorization, deep learning, and beyond.
\newblock {\em arXiv preprint arXiv:1506.07540}, 2015.

\bibitem{sagun2017empirical}
Levent Sagun, Utku Evci, V~Ugur Guney, Yann Dauphin, and Leon Bottou.
\newblock Empirical analysis of the hessian of over-parametrized neural
  networks.
\newblock {\em arXiv preprint arXiv:1706.04454}, 2017.

\bibitem{shalev2017failures}
Shai Shalev-Shwartz, Ohad Shamir, and Shaked Shammah.
\newblock Failures of gradient-based deep learning.
\newblock {\em arXiv preprint arXiv:1703.07950}, 2017.

\bibitem{safran2016quality}
Itay Safran and Ohad Shamir.
\newblock On the quality of the initial basin in overspecified neural networks.
\newblock In {\em International Conference on Machine Learning}, pages
  774--782, 2016.

\bibitem{saxe2013exact}
Andrew~M Saxe, James~L McClelland, and Surya Ganguli.
\newblock Exact solutions to the nonlinear dynamics of learning in deep linear
  neural networks.
\newblock {\em arXiv preprint arXiv:1312.6120}, 2013.

\bibitem{arora2014provable}
Sanjeev Arora, Aditya Bhaskara, Rong Ge, and Tengyu Ma.
\newblock Provable bounds for learning some deep representations.
\newblock In {\em International Conference on Machine Learning}, pages
  584--592, 2014.

\bibitem{pascanu2014saddle}
Razvan Pascanu, Yann~N Dauphin, Surya Ganguli, and Yoshua Bengio.
\newblock On the saddle point problem for non-convex optimization.
\newblock {\em arXiv preprint arXiv:1405.4604}, 2014.

\bibitem{dauphin2014identifying}
Yann~N Dauphin, Razvan Pascanu, Caglar Gulcehre, Kyunghyun Cho, Surya Ganguli,
  and Yoshua Bengio.
\newblock Identifying and attacking the saddle point problem in
  high-dimensional non-convex optimization.
\newblock In {\em Advances in neural information processing systems}, pages
  2933--2941, 2014.

\bibitem{arora2016provable}
Sanjeev Arora, Rong Ge, Tengyu Ma, and Andrej Risteski.
\newblock Provable learning of noisy-or networks.
\newblock {\em arXiv preprint arXiv:1612.08795}, 2016.

\bibitem{arora2018optimization}
Sanjeev Arora, Nadav Cohen, and Elad Hazan.
\newblock On the optimization of deep networks: Implicit acceleration by
  overparameterization.
\newblock {\em arXiv preprint arXiv:1802.06509}, 2018.

\bibitem{kidambi2018insufficiency}
Rahul Kidambi, Praneeth Netrapalli, Prateek Jain, and Sham~M Kakade.
\newblock On the insufficiency of existing momentum schemes for stochastic
  optimization.
\newblock {\em arXiv preprint arXiv:1803.05591}, 2018.

\bibitem{gal2015dropout}
Yarin Gal and Zoubin Ghahramani.
\newblock Dropout as a bayesian approximation: Insights and applications.
\newblock In {\em Deep Learning Workshop, ICML}, volume~1, page~2, 2015.

\bibitem{arora2017generalization}
Sanjeev Arora, Rong Ge, Yingyu Liang, Tengyu Ma, and Yi~Zhang.
\newblock Generalization and equilibrium in generative adversarial nets (gans).
\newblock {\em arXiv preprint arXiv:1703.00573}, 2017.

\bibitem{arora2017gans}
Sanjeev Arora and Yi~Zhang.
\newblock Do gans actually learn the distribution? an empirical study.
\newblock {\em arXiv preprint arXiv:1706.08224}, 2017.

\bibitem{levine2017benefits}
Yoav Levine, Or~Sharir, and Amnon Shashua.
\newblock Benefits of depth for long-term memory of recurrent networks.
\newblock {\em arXiv preprint arXiv:1710.09431}, 2017.

\bibitem{rorschach1922psychodiagnostik}
Hermann Rorschach.
\newblock Psychodiagnostik.
\newblock {\em The Journal of Nervous and Mental Disease}, 56(3):306, 1922.

\bibitem{haykin2008adaptive}
Simon~S Haykin.
\newblock {\em Adaptive filter theory}.
\newblock Pearson Education India, 2008.

\bibitem{barker1995bayesian}
Allen~L Barker, Donald~E Brown, and Worthy~N Martin.
\newblock Bayesian estimation and the kalman filter.
\newblock {\em Computers \& Mathematics with Applications}, 30(10):55--77,
  1995.

\bibitem{charles2015dynamics}
Adam~Shabti Charles.
\newblock {\em Dynamics and correlations in sparse signal acquisition}.
\newblock PhD thesis, Georgia Institute of Technology, 2015.

\bibitem{moon2000mathematical}
Todd~K Moon and Wynn~C Stirling.
\newblock {\em Mathematical methods and algorithms for signal processing},
  volume~1.
\newblock Prentice hall New York, 2000.

\bibitem{saxe2018iclr}
Andrew~Michael Saxe, Yamini Bansal, Joel Dapello, Madhu Advani, Artemy
  Kolchinsky, Brendan~Daniel Tracey, and David~Daniel Cox.
\newblock On the information bottleneck theory of deep learning.
\newblock {\em International Conference on Learning Representations}, 2018.

\bibitem{zhang2016understanding}
Chiyuan Zhang, Samy Bengio, Moritz Hardt, Benjamin Recht, and Oriol Vinyals.
\newblock Understanding deep learning requires rethinking generalization.
\newblock {\em ICLR}, 2017.

\bibitem{ghorbani2017interpretation}
Amirata Ghorbani, Abubakar Abid, and James Zou.
\newblock Interpretation of neural networks is fragile.
\newblock {\em arXiv preprint arXiv:1710.10547}, 2017.

\bibitem{rozell2008sparse}
Christopher~J Rozell, Don~H Johnson, Richard~G Baraniuk, and Bruno~A Olshausen.
\newblock Sparse coding via thresholding and local competition in neural
  circuits.
\newblock {\em Neural computation}, 20(10):2526--2563, 2008.

\bibitem{hu2012network}
Tao Hu, Alexander Genkin, and Dmitri~B Chklovskii.
\newblock A network of spiking neurons for computing sparse representations in
  an energy-efficient way.
\newblock {\em Neural computation}, 24(11):2852--2872, 2012.

\bibitem{charles2012common}
Adam~S Charles, Pierre Garrigues, and Christopher~J Rozell.
\newblock A common network architecture efficiently implements a variety of
  sparsity-based inference problems.
\newblock {\em Neural computation}, 24(12):3317--3339, 2012.

\bibitem{sussillo2013opening}
David Sussillo and Omri Barak.
\newblock Opening the black box: low-dimensional dynamics in high-dimensional
  recurrent neural networks.
\newblock {\em Neural computation}, 25(3):626--649, 2013.

\bibitem{rivkind2017local}
Alexander Rivkind and Omri Barak.
\newblock Local dynamics in trained recurrent neural networks.
\newblock {\em Physical review letters}, 118(25):258101, 2017.

\bibitem{rajan2006eigenvalue}
Kanaka Rajan and LF~Abbott.
\newblock Eigenvalue spectra of random matrices for neural networks.
\newblock {\em Physical review letters}, 97(18):188104, 2006.

\bibitem{khrulkov2017expressive}
Valentin Khrulkov, Alexander Novikov, and Ivan Oseledets.
\newblock Expressive power of recurrent neural networks.
\newblock {\em arXiv preprint arXiv:1711.00811}, 2017.

\bibitem{jaeger2001short}
Herbert Jaeger.
\newblock {\em Short term memory in echo state networks}, volume~5.
\newblock GMD-Forschungszentrum Informationstechnik, 2001.

\bibitem{buonomano2009state}
Dean~V Buonomano and Wolfgang Maass.
\newblock State-dependent computations: spatiotemporal processing in cortical
  networks.
\newblock {\em Nature Reviews Neuroscience}, 10(2):113, 2009.

\bibitem{charles2014short}
Adam~S Charles, Han~Lun Yap, and Christopher~J Rozell.
\newblock Short-term memory capacity in networks via the restricted isometry
  property.
\newblock {\em Neural computation}, 26(6):1198--1235, 2014.

\bibitem{maass2002real}
Wolfgang Maass, Thomas Natschl{\"a}ger, and Henry Markram.
\newblock Real-time computing without stable states: A new framework for neural
  computation based on perturbations.
\newblock {\em Neural computation}, 14(11):2531--2560, 2002.

\bibitem{charles2017distributed}
Adam~S Charles, Dong Yin, and Christopher~J Rozell.
\newblock Distributed sequence memory of multidimensional inputs in recurrent
  networks.
\newblock {\em Journal of Machine Learning Research}, 18(7):1--37, 2017.

\bibitem{zhu2013visual}
Mengchen Zhu and Christopher~J Rozell.
\newblock Visual nonclassical receptive field effects emerge from sparse coding
  in a dynamical system.
\newblock {\em PLoS computational biology}, 9(8):e1003191, 2013.

\bibitem{sussillo2015neural}
David Sussillo, Mark~M Churchland, Matthew~T Kaufman, and Krishna~V Shenoy.
\newblock A neural network that finds a naturalistic solution for the
  production of muscle activity.
\newblock {\em Nature neuroscience}, 18(7):1025, 2015.

\bibitem{depasquale2018full}
Brian DePasquale, Christopher~J Cueva, Kanaka Rajan, LF~Abbott, et~al.
\newblock full-force: A target-based method for training recurrent networks.
\newblock {\em PloS one}, 13(2):e0191527, 2018.

\bibitem{rajan2016recurrent}
Kanaka Rajan, Christopher~D Harvey, and David~W Tank.
\newblock Recurrent network models of sequence generation and memory.
\newblock {\em Neuron}, 90(1):128--142, 2016.

\end{thebibliography}

\end{document}